\documentclass{bmvc2k}

\title{Channel-Partitioned Windowed Attention And Frequency Learning for Single Image Super-Resolution}

\addauthor{Dinh Phu Tran}{phutx2000@kaist.ac.kr}{1}
\addauthor{Dao Duy Hung}{hicehehe@kaist.ac.kr}{1}
\addauthor{Daeyoung Kim}{kimd@kaist.ac.kr}{1}

\addinstitution{
 Korea Advanced Institute of\\
 Science and Technology\\
 Daejeon, Korea
}

\runninghead{TRAN, HUNG, KIM}{Attention \& Frequency Learning for SISR}

\def\eg{\emph{e.g}\bmvaOneDot}

\def\etal{\emph{et al}\bmvaOneDot}

\usepackage{indentfirst}
\usepackage{booktabs}
\usepackage{multirow}
\usepackage{colortbl}
\usepackage{graphicx,wrapfig,lipsum}
\usepackage[export]{adjustbox}
\usepackage{tabto}
\usepackage{amssymb}
\usepackage{float}
\usepackage{amsmath}
\newcommand{\varA}[1]{{\operatorname{#1}}}
\usepackage{caption}

\newcommand{\squeezeup}{\vspace{-2.5mm}}

\begin{document}

\maketitle

\vspace{-3mm}
\begin{abstract}
Recently, window-based attention methods have shown great potential for computer vision tasks, particularly in Single Image Super-Resolution (SISR). However, it may fall short in capturing long-range dependencies and relationships between distant tokens. Additionally, we find that learning on spatial domain does not convey the frequency content of the image, which is a crucial aspect in SISR. To tackle these issues, we propose a new Channel-Partitioned Attention Transformer (CPAT) to better capture long-range dependencies by sequentially expanding windows along the height and width of feature maps. In addition, we propose a novel Spatial-Frequency Interaction Module (SFIM), which incorporates information from spatial and frequency domains to provide a more comprehensive information from feature maps. This includes information about the frequency content and enhances the receptive field across the entire image. Experimental findings show the effectiveness of our proposed modules and architecture. In particular, CPAT surpasses current state-of-the-art methods HAT by up to \textbf{0.31dB} at x2 SR on Urban100.
\end{abstract}

\section{Introduction}
\label{sec:intro}
Single Image Super-Resolution (SISR) is a low-level vision task that aims to enhance a low-resolution (LR) image into a high-resolution (HR) image.

Initially, convolutional neural networks~\cite{dong2015image,kim2016accurate, kim2016deeply, zhang2018image, Li_2018_ECCV} achieved outstanding results in SISR a few years ago. However, it does not possess the capability to gather global contextual information, as they mainly focus on nearby areas and might miss out on important connections that are far apart. Recently, Transformers, which utilize self-attention mechanisms, excel at modeling long-range dependencies not only with high-level vision tasks such as Image Captioning~\cite{wang2022end, he2020image}, 3D-aware Image Synthesis~\cite{sargent2023vq3d}, $etc.$, but also with low-level vision tasks such as localization~\cite{10.1007/978-3-031-20080-9_41}, segmentation~\cite{9133304, chen2021transunet, TranNPT22}, $etc.$, including SISR~\cite{chen2021pre, liang2021swinir, chen2022cross, chen2023activating}. 

Although Transformer has shown great performance in SISR compared to CNN-based methods, they still have limitations that need to be dealt with. The use of dense attention in IPT~\cite{chen2021pre} focuses on short token sequences that come from a dense area of an image. As a result, the receptive field is restricted due to this approach. SwinIR~\cite{liang2021swinir} employs a Swin Transformer as the main backbone, which has a main drawback is that it limits the receptive fields for extracting information from global information. Although HAT~\cite{chen2023activating} is state-of-the-art in SISR, it still uses Swin Transformer, thus limiting the extraction of global information. Therefore, the current methods still can not fully exploit the potential of Transformer for SISR. On the other hand, current methods in SISR mainly extract features from the spatial domain without leveraging features extracted from the frequency domain, which include valuable information and are beneficial for HR image reconstruction.

In order to tackle the mentioned drawbacks and unlock more potential of Transformer for SISR, we propose a novel architecture named Channel-Partitioned Attention Transformer (CPAT), depicted in Fig.~\ref{fig:overall_arch}. A key component in our CPAT is the new self-attention mechanism called \textbf{Channel-Partitioned Windowed Self-Attention (CPWin-SA)} to better capture long-range information and relationships between distant tokens. In addition, we also design a \textbf{Spatial-Frequency Interaction Module (SFIM)} to integrate the spatial and frequency domains to fully exploit the information from feature maps, thereby boosting the quality of output images. Based on these designs, our method can extract robust features to aid in reconstruction and achieve significant improvements compared to the current methods.

\textbf{Contributions:} \textbf{1)} We propose a novel Channel-Partitioned Windowed Self-Attention (CPWin-SA), a robust  feature extraction for better reconstruction of images. \textbf{2)}. We design a new Spatial-Frequency Interaction Module
(SFIM) to leverage all features from both spatial and frequency domains that improve the model's performance. \textbf{3)} Our network outperforms the current state-of-the-art methods for SISR.

\vspace{-4mm}
\section{Related Work}
\label{sec:related_work}

\textbf{Deep Neural Networks for SISR}. Dong \etal~\cite{dong2015image} conducted the first study utilizing deep learning in SISR, called SRCNN, a simple yet effective three-layer CNN for SISR. Following SRCNN, CNNs have been employed in subsequent studies to enhance SISR performance~\cite{lim2017enhanced, ledig2017photo, zhang2018residual, kim2016deeply}. Recently, Transformer has been used in various computer vision applications, from localization~\cite{rambhatla2023most, zhao2018weakly} to deeply understanding images~\cite{jaderberg2015spatial, sun2022locate, zhu2022transgeo, shi2022motion} to sequence-based networks~\cite{bluche2016joint, miech17loupe, Grechishnikova863415}. ViT~\cite{dosovitskiy2020vit} was proposed by Dosovitskiy \etal that processes input images by segmenting them into patches and then projecting these patches into sequential tokens as input of transformer module and achieved remarkable results with high-level vision tasks. To reduce the high computational cost of ViT, Liu \etal~\cite{liu2021swin} proposed a hierarchical transformer called Swin Transformer, using self-attention over local windows instead of the entire image like ViT. In the field of low-level vision tasks, such as SISR, Transformer can also be used as a powerful backbone. EDT ~\cite{li2021efficient} performs pre-training to boost low-level task. IPT~\cite{chen2021pre} leveraged the pre-trained transformer to improve performance for the image super-resolution task. GRL~\cite{li2023efficient} proposed the anchored stripe self-attention to reach the modelling capacity beyond the regional range for image restoration. SwinIR~\cite{liang2021swinir} utilized Swin Transformer as a deep feature extractor and achieved impressive results for image restoration. HAT~\cite{chen2023activating} proposed Hybrid Attention Transformer and Overlapping-cross Attention, achieving state-of-the-art SISR performance. However, the common limitation of these works is that they are limited in capturing long-range dependencies and may miss the connection with the distant tokens. Our proposed Transformer can handle this problem by enhancing window size in the window-based attention mechanism while still being efficient for high-resolution images.

\noindent
\textbf{Frequency domain in Computer Vision.} The frequency domain is widely used in digital signal processing~\cite{trider1978fast, saxena2005fractional, kim2001signal} and benefits the computer vision domain~\cite{he2016salient, 7217088, Cai2021FreqNetAF, Wang_2023_CVPR}, such as image super-resolution. Cai \etal proposed FreqNet~\cite{Cai2021FreqNetAF} consisting of two main branches: spatial and frequency branches. FreqNet transforms LR and HR images to the frequency domain using discrete cosine transform (DCT)~\cite{1672377}. DCT features combine with spatial features, and then, the inverse DCT (iDCT)~\cite{1672377} is used to convert the feature maps back to the spatial domain. FreqNet uses a dual branch and DCT transform from the beginning, which increases the computational complexity. Wang \etal proposed SFMNet~\cite{Wang_2023_CVPR} for face super-resolution (FSR). SFMNet uses Fourier transform~\cite{brigham1988fast} in the frequency domain to convert spatial features to the frequency domain to capture the global facial structure and inverse Fourier transform~\cite{brigham1988fast} to convert frequency features to the spatial domain. Similar to FreqNet, the frequency branch in SFMNet is too complex and the computational complexity is high due to it computes spatial-frequency cross-attention module multiple times to combine spatial and frequency domains. Our Spatial-Frequency Integrated Module (SFIM) is designed to deal with this problem. It is simple yet effectively leverages features (textures, edges, $etc.$) from the frequency domain that may miss extraction in the spatial domain while the computational complexity is not significantly increasing.

\vspace{-3.5mm}
\section{Methodology}
\label{sec:methodology}
\label{sec:net_arch}
\subsection{The Overall Architecture}

\begin{figure*}
\begin{center}
\scalebox{0.9}{
\includegraphics[width=1.\textwidth]{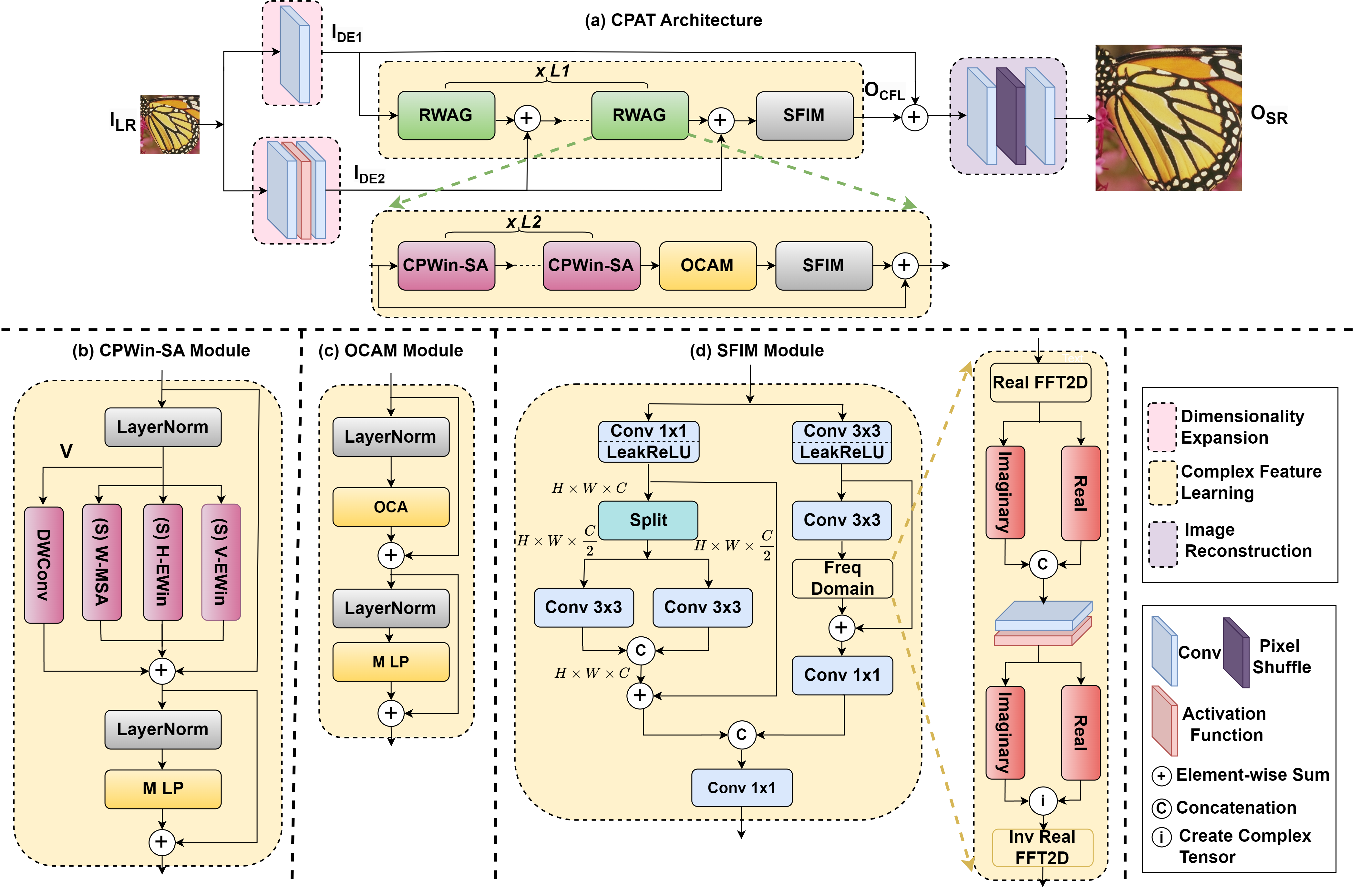}}
\end{center}
   \caption{Architecture details. (a) The overall architecture of CPAT. (b) Structure of Channel-Partitioned Windowed Self-Attention. (c) Structure of Overlapping Cross-Attention Module. (d) Spatial-Frequency Integrated Module.}
\label{fig:overall_arch}
\vspace{-5.8mm}
\end{figure*}
The overall architecture of Channel-Partitioned Attention Transformer (CPAT) is shown in Fig.~\ref{fig:overall_arch}(a), which consists of three parts: Dimensionality Expansion (DE), Complex Feature Learning (CFL), and Image Reconstruction (IR). We set the input LR and output SR images as $I_{LR} \in \mathbb{R}^{H\times W\times C_{in}}$ and $O_{SR}$ $\in \mathbb{R}^{H\times W\times C_{out}}$, where $C_{in}$ and $C_{out}$ are the channel numbers of the LR and SR images, respectively. First, Dimensionality Expansion transforms $I_{LR}$ from a low-dimensional space to a high-dimensional space $I_{DE1}, \;I_{DE2}$ as follows:
\begin{equation} \label{e:equation1} 
I_{DE1} = F_{c3}(I_{LR}), \;I_{DE2} = F_{c2}(I_{LR}),
\end{equation} where $I_{DE1}$ and $I_{DE2}$ are the output of DE; $F_{c3}$ is a 3x3 conv, and $F_{c2}$ is a convolution stage (Convolution - Activation - Convolution). Next, $I_{DE1}$ goes through Complex Feature Learning, which involves a series of RWAGs to learn the complex and deep features. Each RWAG consists of several CPWin-SA modules, an overlapping cross-attention module (OCAM), and a SFIM module.
The output of CFL $O_{CFL}$ is 
\begin{equation} \label{e:equation3}
\setlength\abovedisplayskip{4.0pt}
O_{CFL} = F_{SFIM}(F_{RWAG}^{L_{1}}(F_{RWAG}^{L_{1}-1}(...(F_{RWAG}^{1}(I_{DE1})+I_{DE2}))+I_{DE2})+I_{DE2}),
\end{equation} 
where $F_{RWAG}^{i}, \;F_{SFIM}$ represent the functions of i-th RWAG, SFIM modules, $i = 1, 2, ..., L_{1}$, and $L_{1}$ is the number of RWAG in CFL.
Finally, $O_{CFL}$ is passed to the Image Reconstruction (IR) to obtain the output $O_{SR}$. IR includes convolution layers and PixelShuffle~\cite{7780576}. We set the function of IR as $F_{IR}$, and we then have
\begin{equation} \label{e:equation4}
O_{SR} = F_{IR}(O_{CFL} + I_{DE1}),
\end{equation}

\vspace{-6mm}
\subsection{Channel-Partitioned Windowed Self-Attention (CPWin-SA)}

\vspace{-4mm}
\begin{figure*}[h!]
\centering
\includegraphics[width=1\textwidth]{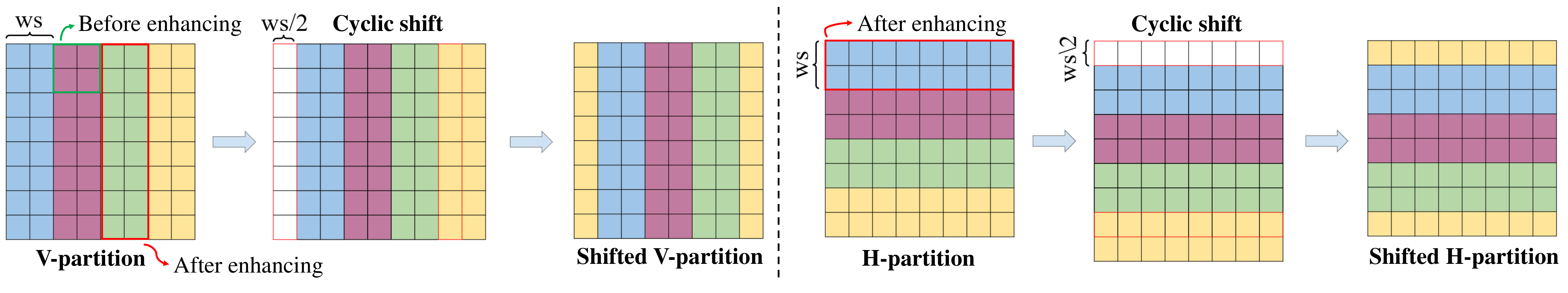}
    \vspace*{-2mm}
   \caption{Enhanced window strategy and One-Direction Shift Operation in V-EWin and H-EWin}
\label{fig:shifted_operation}
\squeezeup
\end{figure*}

Channel-Partitioned Windowed Self-Attention (CPWin-SA) is a key component of our method, described in Fig.~\ref{fig:overall_arch}(b). It consists of three different attention mechanism types: Vertically Enhanced Window Attention (H-EWin), Horizontally Enhanced Window Attention (H-EWin), and standard Windowed Multi-head Self-Attention (W-MSA). We split the input feature maps along the channel dimension into three equal parts, corresponding to the three attentions that are above mentioned.

\noindent
\textbf{Enhanced Window Self-Attention.} This refers to V-EWin and H-EWin. We extend the squared windows along the vertical and horizontal directions of the input feature maps, shown in Fig.~\ref{fig:shifted_operation}. Specifically, the window size is extended from $ws\times ws$ (squared window) to $H\times ws$ (V-EWin) or $ws\times W$ (H-EWin), where $ws$ is the window size ($ws < H$, $ws < W$). V-EWin and H-Ewin enhance attention areas by extending the window size, thereby increasing the ability to extract global contextual information and relationships between distant tokens. For simplicity, the mathematical descriptions below are used for V-EWin, and similar formulations can be applied to H-EWin.
With the input feature $X \in \mathbb{R}^{H\times W\times \frac{C}{3}}$ (after splitting), we compute attention $N$ times in parallel, where $N$ is the head number. We partition X into non-overlapping windows of size $H\times ws$ for each attention head, then calculate the self-attention of the $i$-th window feature as $X_{i} \in \mathbb{R}^{H\times ws\times \frac{C}{3}}$, $i$=1,...,$\frac{H\times W}{H\times ws}$  for the $n$-th head,
\begin{equation} \label{e:equation5}
\begin{array}{l}
Y_{i}^n = Attention(Q_{i}^n,\; K_{i}^n,\; V_{i}^n) = Attention(X_{i}W_{n}^Q,\; X_{i}W_{n}^K,\; X_{i}W_{n}^V),
\end{array}
\end{equation}
\noindent
where $Y_{i}^n \in \mathbb{R}^{H\times ws\times D}$ denotes the attention output of $X_{i}$ in the $n$-th head, $D$=$\frac{C}{3N}$ represents the channel dimension in each head. $Q_{i}^n,\; K_{i}^n,\; V_{i}^n \in \mathbb{R}^{\frac{C}{3}\times D}$ are the projection matrices of query, key, and value, respectively for $n$-th head. We use conditional position embedding (CPE)~\cite{chu2023CPVT} to add the spatial relationships into the network. The attention feature $Y^n \in \mathbb{R}^{H\times W\times D}$ of $X$ is obtained by calculating the attention operation on all $X_{i}$, then performing reshaping and merging them in the order of division. We then concatenate the output of all heads and combine them with a final weight matrix to achieve the attention map output of V-EWin.
\begin{equation} \label{e:equation6}
V\varA{-}EWin(x) = Concat(Y^{1},..., Y^{N-1}, Y^{N})W^p,
\end{equation}
\noindent
where the $W^p \in \mathbb{R}^{\frac{C}{3}\times \frac{C}{3}}$ denotes the projection matrix for feature fusion.

\noindent
\textbf{One-Direction Shift Operation.} The window-based self-attention module causes a lack of information linkage between windows, thereby reducing the modeling power with distant tokens. We enhance the windows along the height and width of the feature map, we propose One-Directional Shift Operation instead of two directions as in Swin Transformer while ensuring the transformer's modeling power, detailed in Fig.~\ref{fig:shifted_operation}. For V-EWin, we move the windows to the left by a distance of $\frac{ws}{2}$ pixels, while H-EWin moves downward by also $\frac{ws}{2}$ pixels. Then, we use a cyclic shift to complete the shift operation. After computing the attention operation on the shifted feature map, we then revert this feature map to obtain the feature map, which is in the original order.

\noindent
\textbf{Squared Window Self-Attention.} For Squared Window Self-Attention, we utilize Swin Transformer (W-MSA)~\cite{liu2021swin}. W-MSA is computed entirely similarly to the Enhanced Window Self-Attention we presented above. The only difference lies in the window size being $ws\times ws$ instead of $H\times ws$ or $ws\times W$. Using squared windows helps to focus on local features. Combining standard (squared) and enhanced windows and computing self-attention in various window shapes benefits datasets that contain many texture features in various directions. 

\noindent
\textbf{Channel-Partitioned Windowed Self-Attention.} CPWin-SA consists of three attention modules (V-EWin, H-EWin, and W-MSA) and an MLP, which includes a GELU~\cite{hendrycks2016gelu} activation function between 2 linear projection layers. Because Transformer has no inductive bias, we simply use a depthwise convolution to add inductive bias that aims to improve the performance of Transformer. A Layer Norm layer is used before the attention modules and MLP. The entire process of CPWin-SA is as follows:
\begin{subequations}
\begin{align}
\begin{split}
X = LayerNorm(X_{in}),\; c = C/3, \label{e:equation7a}
\end{split}\\
\begin{split}
X_{1} = V\varA{-}EWin(X[:,:,:c]),\;
X_{2} = H\varA{-}EWin(X[:,:,c:2c]),\;
X_{3} = W\varA{-}MSA(X[:,:,2c:]), \label{e:equation7b}
\end{split}\\
\begin{split}
\hat{X} = Concat(X_{1},\; X_{2},\; X_{3}) + X_{in} + DWConv(V), \label{e:equation7c}
\end{split}\\
\begin{split}
X_{out} = MLP(LayerNorm(\hat{X})) + \hat{X},\label{e:equation7d}
\end{split}
\end{align}
\end{subequations}

\noindent
where $X_{in}$, $X_{out}$ and $C$ are CPWin-SA's input, output features, and the channel numbers; DWConv and V are depthwise convolution and $value$ matrix. Shift Operation uses two consecutive Transformer modules to increase the interaction among non-overlapping windows. The computational complexity of global MSA (self-attention is computed on the full feature map) and V-EWin are
\begin{subequations}
\begin{align}
\begin{split}
\mathcal{O}(Global-MSA) = 4HW(C/3)^2 + 2(HW)^2(C/3)), \label{e:equation8a}
\end{split}\\
\begin{split}
\mathcal{O}(V\varA{-}EWin) = 4HW(C/3)^2 + 2H^{2}Wws(C/3)), \label{e:equation8b}
\end{split}
\end{align}
\end{subequations}
Assuming H=W (squared image) and $ws\ll H,\; C\ll H$, the computational complexity of $V\varA{-}EWin$ is $\mathcal{O}(H^2Wws(C/3))$ = $\mathcal{O}(H^3)$ whereas $Global-MSA$ is $\mathcal{O}((HW)^2(C/3))$ = $\mathcal{O}(H^4)$. Therefore, our proposed Transformer can be applied to high-resolution input images.

\vspace{-2mm}
\subsection{Overlapping Cross-Attention Module (OCAM)}



OCAM~\cite{chen2023activating} enhances the connections between windows by partitioning feature maps into overlapping windows and calculating self-attention on each window to improve the performance of the network. The structure of OCAM is depicted in Fig.~\ref{fig:overall_arch}(c). Specifically, for $X_{Q}$, $X_{K}$, $X_{v}$ from the input feature $X$, $X_{Q}$ is divided
into non-overlapping windows of size $M\times M$, with $\frac{HW}{M^2}$ being the total number of windows. $X_{K}$ and $X_{V}$ are unfolded to $\frac{HW}{M^2}$ overlapping windows of size $M_{o}\times M_{o}$ ($M_{o} > M$),
\begin{equation} \label{e:equation9}
M_{o} = (1 + \alpha)\times M,
\vspace{-2mm}
\end{equation}
where $\alpha$ is overlapping ratio. Then, self-attention is computed in each windows as in Eq.~\ref{e:equation5}.

\vspace{-2mm}
\subsection{Spatial-Frequency Interaction Module (SFIM)}
Spatial features may lack frequency information and fine-grained details that are important for HR image reconstruction. To address these issues, we carefully design a Spatial-Frequency Interaction Module (SFIM) shown in Fig.~\ref{fig:overall_arch}(d) to leverage the spatial and frequency domain features. SFIM consists of two branches: spatial and frequency branches. Spatial branch is helpful in extracting local spatial features. We denote the input, output of SFIM, and output of spatial as
$I_{SFIM}$, $O_{SFIM}$, $O_{SB}$, respectively, then spatial branch is represented as follows:
\begin{subequations}
\begin{align}
\begin{split}
O_{SB1} = F_{A}(F_{c1}(I_{SFIM})), \label{e:equation10a}
\end{split}\\
\begin{split}
O_{SB} = Concat(F_{c3}(O_{SB1}[:, :C/2, :, :]), \;F_{c3}(O_{SB1}[:, C/2:, :, :])) + O_{SB1}, \label{e:equation10b}
\end{split}
\end{align}
\end{subequations} where $F_{c1}$, $F_{c3}$, $F_{A}$ and $C$  are 1x1 conv, 3x3 conv, LeakReLU~\cite{maas2013rectifier}, and number channel, respectively. Frequency branch is used for capturing global structure and frequency information.
To convert spatial features into the frequency domain, we utilize the Fast Fourier Transform (FFT)~\cite{cooley1965algorithm}, and use inverse FFT (iFFT)~\cite{cooley1965algorithm} to convert frequency features back into the spatial domain. The space in which the Fourier transform is performed contains complex orthonormal basis functions. These basis functions are represented by complex frequency components $X \in \mathbb{C}^{U\times V\times C}$, which describe the following equation.

\begin{equation} \label{e:equation_add1}
\mathcal{F}\{x\}_{u,v} = X_{u,v} = \frac{1}{\sqrt{HW}} \sum_{h=0}^{H-1} \sum_{w=0}^{W-1} x_{h,w} e^{-i2\pi \left( u\frac{h}{H} + v\frac{w}{W} \right)},
\end{equation}

where $x_{h, w}$ is the pixel value; $(u, v)$ represents the coordinate of a spatial frequency on the frequency spectrum; $\mathcal{F}\{x\}_{u,v}$ is the complex frequency value; $e$ and $i$ are Euler’s number and the imaginary unit, respectively. The FFT capture global patterns, and high-frequency information in an image that might be less apparent in the spatial domain. We denote the output of frequency branch as $O_{FB}$, then frequency branch can be represented as follows:
\begin{subequations}
\begin{align}
\begin{split}
O_{FB1} = F_{A}(F_{c3}(I_{SFIM})), \label{e:equation11a}
\end{split}\\
\begin{split}
O_{FB} = F_{c1}(F_{FD}(F_{c3}(O_{FB1})) + O_{FB1}), \label{e:equation11b}
\end{split}
\end{align}
\end{subequations} where $F_{FD}$ is $Freq \;Domain$, which is the key components of SFIM helps the model pay more attention to high-frequency information. At the end, we combine the outputs of spatial and frequency domains to obtain the output feature of SFIM is $O_{SFIM}$
\begin{equation} \label{e:equation12}
O_{SFIM} = F_{c1}(Concat([O_{SB}, O_{FB}])),
\end{equation}

\section{Experiments}
\label{sec:exps}
\subsection{Experimental Settings}

We use DF2K (DIV2K~\cite{8014884}+Flicker2K~\cite{lim2017enhanced}) dataset as the training set for a fair comparison with other methods. For evaluating our model, we use five benchmark datasets: Set5~\cite{Bevilacqua2012LowComplexitySS}, Set14~\cite{inproceedings}, BSD100~\cite{937655}, Urban100~\cite{7299156}, and Manga109~\cite{Matsui_2016}. Furthermore, we use peak signal-to-noise ratio (PSNR)~\cite{wang2004image} and the structural similarity index measure (SSIM)~\cite{wang2004image} as quantitative metrics, which are calculated on the Y channel in YCbCr Space.
For the structure of network, we set the number of RWAG and SPWin-SA to 6, the channel number to 180, and the window size is set to  16. The overlapping ratio in OCAM remains at 0.5 as in~\cite{chen2023activating}. We apply the self-ensemble strategy similarly to~\cite{timofte2016seven} in testing that we call CPAT† as in~\cite{zhang2018image, liang2021swinir, chen2023dual}.
We use a patch size of $64\times 64$ and a batch size of 32 during the training. We simply use L1 loss, and Adam optimizer~\cite{Kingma2014AdamAM} to optimize models with 500K iterations.

\vspace{-2mm}
\subsection{Ablation Study}
\vspace{-1mm}
Following~\cite{chen2022cross, chen2023dual}, we train x2 SR model on DF2K (DIV2K~\cite{8014884}+Flicker2K~\cite{lim2017enhanced}), and test on Urban100~\cite{7299156} for all experiments in this section. FLOPs are calculated on a 256x256 HR image. Results are reported in Tab.~\ref{wrap-tab:1},~\ref{wrap-tab:2},~\ref{wrap-tab:3},~\ref{wrap-tab:4}, and the better results are shown in \textbf{bold}.

\begin{table}[!t]
    \begin{minipage}{0.345\textwidth}
    \setlength{\tabcolsep}{0.50mm}
    \scalebox{0.79}{
        \begin{tabular}[t]{|c|c|c|c|}
        \hline
        Structure &PSNR &SSIM & FLOPs\\ \hline \hline
        Squared wins & 34.03 & 0.9438 & 324.8G\\
        \hline
        Enhanced wins & \textbf{34.26} & \textbf{0.9448} & 329.0G\\
        \hline
        \end{tabular}}
        \vspace{3mm}
        \caption{Effect of the enhanced windows.}\label{wrap-tab:1}
    \end{minipage}%
    \hspace{0.001cm}
    \begin{minipage}{0.31\textwidth}
      \setlength{\tabcolsep}{1.0mm}
      \scalebox{0.79}{
        \begin{tabular}[t]{|c|c|c|c|}
\hline
Structure&PSNR&SSIM&FLOPs\\ \hline\hline
w/o shift & 34.13 & 0.9440 & 329.0G\\
\hline
w/ shift & \textbf{34.26} & \textbf{0.9448} & 329.0G\\
\hline
\end{tabular}}
\vspace{3mm}
\caption{Effect of One-Direction Shift Operation.}\label{wrap-tab:2}
    \end{minipage} \hspace{0.01cm}
    \begin{minipage}{0.31\textwidth}
    \setlength{\tabcolsep}{1.0mm}
    \scalebox{0.79}{
    \begin{tabular}[t]{|c|c|c|c|}
\hline
Structure&PSNR&SSIM&FLOPs\\ \hline\hline
w/o SFIM & 34.14 & 0.9442 & 256.2G\\
\hline
w/ SFIM & \textbf{34.26} & \textbf{0.9448} & 329.0G\\
\hline
\end{tabular}}
\vspace{3mm}
\caption{Effect of SFIM module.}\label{wrap-tab:3}
\end{minipage}
\vspace{-4mm}
\end{table}


\noindent
\textbf{Effect of the enhanced window.} Tab.~\ref{wrap-tab:1} shows the effectiveness of the enhanced window in V-EWin and H-EWin instead of the squared window in Swin Transformer. PSNR, when using the enhanced window is 34.26dB compared to 34.03dB with the squared window. These results show that the standard window-based transformer has yet to fully utilize its potential for SISR, whereas our method shows significant improvement and 
the computation complexity does not increase much (FLOPs is 329.0G compared to 324.8G). These results also indicate that transformer-based methods remain a promising research direction in SISR.

\noindent
\textbf{Effect of shift operation.} Tab.~\ref{wrap-tab:2} shows the effect of One-Direction Shifted Operation. When One-Direction Shift Operation is applied, PSNR value is 34.26dB, which is higher compared to 34.13dB without this operation. Additionally, SSIM also increases from 0.9440 to 0.9448. Our shift operation is helpful in getting correlation attention between different window partitions, thereby enhancing the performance of CPWin-SA and our network in general.

\begin{wraptable}{r}{0.356\textwidth}
\vspace{-4.2mm}
\setlength{\tabcolsep}{0.3mm}
\scalebox{0.79}{
\begin{tabular}[t]{|c|c|c|c|}
\hline
Structure&PSNR&SSIM&FLOPs\\ \hline\hline
w/o Freq Domain & 34.03 & 0.9434 & 321.4G\\
\hline
w/ Freq Domain & \textbf{34.26} & 0.9448 & 329.0G\\
\hline
\end{tabular}}
\vspace{2.6mm}
\caption{Effect of Freq Domain module in SFIM}\label{wrap-tab:4}
\end{wraptable}

\noindent
\textbf{Effect of SFIM.} Tab.~\ref{wrap-tab:3} shows the effectiveness of SFIM. We replace SFIM module with a 3x3 conv (we call it "w/o SFIM") and see how it affects PSNR/SSIM. PSNR when having SFIM is 34.26dB, while PSNR when not having SFIM is 34.14dB. By leveraging information from the frequency domain, SFIM boosts the network's performance compared to using convolution only, which provides spatial features.

\noindent
\textbf{Effect of Freq Domain module on SFIM's effectiveness.} To show the effectiveness of frequency features, we conduct an experiment by removing the Freq Domain module from the SFIM module. This means that the SFIM module will solely work with the spatial domain. The results are reported in Tab.~\ref{wrap-tab:4}, which is evident that SFIM operates significantly less effectively when the Freq Domain module is removed. By combining both spatial and frequency domains in the SFIM module, comprehensive features are extracted, thereby enhancing the performance of image reconstruction. Specifically, SFIM with Freq Domain achieves 34.26dB, whereas, without Freq Domain, it achieves 34.03dB. This also demonstrates the potential of combining spatial and frequency domains for other vision tasks.


\begin{table}[h!]
\footnotesize
\setlength{\tabcolsep}{2.55mm}
\centering
\scalebox{0.78}{
\begin{tabular}{|l|c|cc|cc|cc|cc|cc|}
\hline
\multirow{2}{*}{Method} &
  \multirow{2}{*}{Scale} &
  \multicolumn{2}{c|}{Set5~\cite{Bevilacqua2012LowComplexitySS}} &
  \multicolumn{2}{c|}{Set14~\cite{inproceedings}} &
  \multicolumn{2}{c|}{BSD100~\cite{937655}} &
  \multicolumn{2}{c|}{Urban100~\cite{7299156}} &
  \multicolumn{2}{c|}{Manga109~\cite{Matsui_2016}} \\ \cline{3-12} 
 &
   &
  \multicolumn{1}{c|}{PSNR} &
  SSIM &
  \multicolumn{1}{c|}{PSNR} &
  SSIM &
  \multicolumn{1}{c|}{PSNR} &
  SSIM &
  \multicolumn{1}{c|}{PSNR} &
  SSIM &
  \multicolumn{1}{c|}{PSNR} &
  SSIM \\
  \hline\hline
EDSR~\cite{lim2017enhanced} &
  \multirow{12}{*}{x2} &
  \multicolumn{1}{c|}{38.11} &
  0.9601 &
  \multicolumn{1}{c|}{33.92} &
  0.9195 &
  \multicolumn{1}{c|}{32.32} &
  0.9013 &
  \multicolumn{1}{c|}{32.93} &
  0.9351 &
  \multicolumn{1}{c|}{39.10} &
  0.9773 \\
RCAN~\cite{zhang2018image} &
   &
  \multicolumn{1}{c|}{38.27} &
  0.9614 &
  \multicolumn{1}{c|}{34.12} &
  0.9216 &
  \multicolumn{1}{c|}{32.41} &
  0.9027 &
  \multicolumn{1}{c|}{33.34} &
  0.9384 &
  \multicolumn{1}{c|}{39.44} &
  0.9786 \\
NLSA~\cite{9578003} &
   &
  \multicolumn{1}{c|}{38.34} &
  0.9618 &
  \multicolumn{1}{c|}{34.08} &
  0.9231 &
  \multicolumn{1}{c|}{32.43} &
  0.9027 &
  \multicolumn{1}{c|}{33.42} &
  0.9394 &
  \multicolumn{1}{c|}{39.59} &
  0.9789 \\
ELAN~\cite{zhang2022efficient} &
   &
  \multicolumn{1}{c|}{38.36} &
  0.9620 &
  \multicolumn{1}{c|}{34.20} &
  0.9228 &
  \multicolumn{1}{c|}{32.45} &
  0.9030 &
  \multicolumn{1}{c|}{33.44} &
  0.9391 &
  \multicolumn{1}{c|}{39.62} &
  0.9793 \\
IPT$^\ast$~\cite{chen2021pre} &
   &
  \multicolumn{1}{c|}{38.37} &
  - &
  \multicolumn{1}{c|}{34.43} &
  - &
  \multicolumn{1}{c|}{32.48} &
  - &
  \multicolumn{1}{c|}{33.76} &
  - &
  \multicolumn{1}{c|}{-} &
  - \\
RCAN-it~\cite{lin2022revisiting} &
   &
  \multicolumn{1}{c|}{38.37} &
  0.9620 &
  \multicolumn{1}{c|}{34.49} &
  0.9250 &
  \multicolumn{1}{c|}{32.48} &
  0.9034 &
  \multicolumn{1}{c|}{33.62} &
  0.9410 &
  \multicolumn{1}{c|}{39.88} &
  0.9799 \\
SwinIR~\cite{liang2021swinir} &
   &
  \multicolumn{1}{c|}{38.42} &
  0.9623 &
  \multicolumn{1}{c|}{34.46} &
  0.9250 &
  \multicolumn{1}{c|}{32.53} &
  0.9041 &
  \multicolumn{1}{c|}{33.81} &
  0.9427 &
  \multicolumn{1}{c|}{39.92} &
  0.9797 \\
CAT-A~\cite{chen2022cross} &
   &
  \multicolumn{1}{c|}{38.51} &
  0.9626 &
  \multicolumn{1}{c|}{34.78} &
  0.9265 &
  \multicolumn{1}{c|}{32.59} &
  0.9047 &
  \multicolumn{1}{c|}{34.26} &
  0.9440 &
  \multicolumn{1}{c|}{40.10} &
  0.9805 \\
DAT~\cite{chen2023dual} &
   &
  \multicolumn{1}{c|}{38.58} &
  0.9629 &
  \multicolumn{1}{c|}{34.81} &
  0.9272 &
  \multicolumn{1}{c|}{32.61} &
  0.9051 &
  \multicolumn{1}{c|}{34.37} &
  0.9458 &
  \multicolumn{1}{c|}{\color{green}40.33} &
  0.9807 \\
HAT~\cite{chen2023activating} &
   &
  \multicolumn{1}{c|}{\color{green}38.63} &
  \color{green}0.9630 &
  \multicolumn{1}{c|}{\color{green}34.86} &
  \color{green}0.9274 &
  \multicolumn{1}{c|}{\color{green}32.62} &
  \color{green}0.9053 &
  \multicolumn{1}{c|}{\color{green}34.45} &
  \color{green}0.9466 &
  \multicolumn{1}{c|}{40.26} &
  \color{green}0.9809 \\
\textbf{CPAT (Ours)} &
   &
  \multicolumn{1}{c|}{\color{blue}38.68} &
  \color{blue}0.9633 &
  \multicolumn{1}{c|}{\color{blue}34.91} &
  \color{blue}0.9277 &
  \multicolumn{1}{c|}{\color{blue}32.64} &
  \color{blue}0.9056 &
  \multicolumn{1}{c|}{\color{blue}34.76} &
  \color{blue}0.9481 &
  \multicolumn{1}{c|}{\color{blue}40.48} &
  \color{blue}0.9814 \\
\textbf{CPAT† (Ours)} &
   &
  \multicolumn{1}{c|}{\color{red}38.72} &
  \color{red}0.9635 &
  \multicolumn{1}{c|}{\color{red}34.97} &
  \color{red}0.9280 &
  \multicolumn{1}{c|}{\color{red}32.66} &
  \color{red}0.9058 &
  \multicolumn{1}{c|}{\color{red}34.89} &
  \color{red}0.9487 &
  \multicolumn{1}{c|}{\color{red}40.59} &
  \color{red}0.9816 \\ [0.5ex]
  \hline\hline
EDSR~\cite{lim2017enhanced} &
  \multirow{12}{*}{x3} &
  \multicolumn{1}{c|}{34.65} &
  0.9280 &
  \multicolumn{1}{c|}{30.52} &
  0.8462 &
  \multicolumn{1}{c|}{29.25} &
  0.8093 &
  \multicolumn{1}{c|}{28.80} &
  0.8653 &
  \multicolumn{1}{c|}{34.17} &
  0.9476 \\
RCAN~\cite{zhang2018image} &
   &
  \multicolumn{1}{c|}{34.74} &
  0.9299 &
  \multicolumn{1}{c|}{30.65} &
  0.8482 &
  \multicolumn{1}{c|}{29.32} &
  0.8111 &
  \multicolumn{1}{c|}{29.09} &
  0.8702 &
  \multicolumn{1}{c|}{34.44} &
  0.9499 \\
NLSA~\cite{9578003} &
   &
  \multicolumn{1}{c|}{34.85} &
  0.9306 &
  \multicolumn{1}{c|}{30.70} &
  0.8485 &
  \multicolumn{1}{c|}{29.34} &
  0.8117 &
  \multicolumn{1}{c|}{29.25} &
  0.8726 &
  \multicolumn{1}{c|}{34.57} &
  0.9508 \\
ELAN~\cite{zhang2022efficient} &
   &
  \multicolumn{1}{c|}{34.90} &
  0.9313 &
  \multicolumn{1}{c|}{30.80} &
  0.8504 &
  \multicolumn{1}{c|}{29.38} &
  0.8124 &
  \multicolumn{1}{c|}{29.32} &
  0.8745 &
  \multicolumn{1}{c|}{34.73} &
  0.9517 \\
IPT$^\ast$~\cite{chen2021pre} &
   &
  \multicolumn{1}{c|}{34.81} &
  - &
  \multicolumn{1}{c|}{30.85} &
  - &
  \multicolumn{1}{c|}{29.38} &
  - &
  \multicolumn{1}{c|}{29.49} &
  - &
  \multicolumn{1}{c|}{-} &
  - \\
RCAN-it~\cite{lin2022revisiting} &
   &
  \multicolumn{1}{c|}{34.86} &
  0.9308 &
  \multicolumn{1}{c|}{30.76} &
  0.8505 &
  \multicolumn{1}{c|}{29.39} &
  0.8125 &
  \multicolumn{1}{c|}{29.38} &
  0.8755 &
  \multicolumn{1}{c|}{34.92} &
  0.9520 \\
SwinIR~\cite{liang2021swinir} &
   &
  \multicolumn{1}{c|}{34.97} &
  0.9318 &
  \multicolumn{1}{c|}{30.93} &
  0.8534 &
  \multicolumn{1}{c|}{29.46} &
  0.8145 &
  \multicolumn{1}{c|}{29.75} &
  0.8826 &
  \multicolumn{1}{c|}{35.12} &
  0.9537 \\
CAT-A~\cite{chen2022cross} &
   &
  \multicolumn{1}{c|}{35.06} &
  0.9326 &
  \multicolumn{1}{c|}{31.04} &
  0.8538 &
  \multicolumn{1}{c|}{29.52} &
  0.8160 &
  \multicolumn{1}{c|}{30.12} &
  0.8862 &
  \multicolumn{1}{c|}{35.38} &
  0.9546 \\
DAT~\cite{chen2023dual} &
   &
  \multicolumn{1}{c|}{\color{blue}35.16} &
  \color{green}0.9331 &
  \multicolumn{1}{c|}{\color{green}31.11} &
  0.8550 &
  \multicolumn{1}{c|}{\color{green}29.55} &
  \color{green}0.8169 &
  \multicolumn{1}{c|}{30.18} &
  0.8886 &
  \multicolumn{1}{c|}{\color{green}35.59} &
  \color{green}0.9554 \\
HAT~\cite{chen2023activating} &
   &
  \multicolumn{1}{c|}{\color{green}35.07} &
  0.9329 &
  \multicolumn{1}{c|}{31.08} &
  \color{green}0.8555 &
  \multicolumn{1}{c|}{29.54} &
  0.8167 &
  \multicolumn{1}{c|}{\color{green}30.23} &
  \color{green}0.8896 &
  \multicolumn{1}{c|}{35.53} &
  0.9552 \\
\textbf{CPAT (Ours)} &
   &
  \multicolumn{1}{c|}{\color{blue}35.16} &
  \color{blue}0.9334 &
  \multicolumn{1}{c|}{\color{blue}31.15} &
  \color{blue}0.8557 &
  \multicolumn{1}{c|}{\color{blue}29.56} &
  \color{blue}0.8174 &
  \multicolumn{1}{c|}{\color{blue}30.52} &
  \color{blue}0.8923 &
  \multicolumn{1}{c|}{\color{blue}35.66} &
  \color{blue}0.9559 \\
\textbf{CPAT† (Ours)} &
   &
  \multicolumn{1}{c|}{\color{red}35.19} &
  \color{red}0.9335 &
  \multicolumn{1}{c|}{\color{red}31.19} &
  \color{red}0.8559 &
  \multicolumn{1}{c|}{\color{red}29.59} &
  \color{red}0.8177 &
  \multicolumn{1}{c|}{\color{red}30.63} &
  \color{red}0.8934 &
  \multicolumn{1}{c|}{\color{red}35.77} &
  \color{red}0.9563 \\ [0.5ex]
  \hline\hline
EDSR~\cite{lim2017enhanced} &
  \multirow{12}{*}{x4} &
  \multicolumn{1}{c|}{32.46} &
  0.8968 &
  \multicolumn{1}{c|}{28.80} &
  0.7876 &
  \multicolumn{1}{c|}{27.71} &
  0.7420 &
  \multicolumn{1}{c|}{26.64} &
  0.8033 &
  \multicolumn{1}{c|}{31.02} &
  0.9148 \\
RCAN~\cite{zhang2018image} &
   &
  \multicolumn{1}{c|}{32.63} &
  0.9002 &
  \multicolumn{1}{c|}{28.87} &
  0.7889 &
  \multicolumn{1}{c|}{27.77} &
  0.7436 &
  \multicolumn{1}{c|}{26.82} &
  0.8087 &
  \multicolumn{1}{c|}{31.22} &
  0.9173 \\
NLSA~\cite{9578003} &
   &
  \multicolumn{1}{c|}{32.59} &
  0.9000 &
  \multicolumn{1}{c|}{28.87} &
  0.7891 &
  \multicolumn{1}{c|}{27.78} &
  0.7444 &
  \multicolumn{1}{c|}{26.96} &
  0.8109 &
  \multicolumn{1}{c|}{31.27} &
  0.9184 \\
ELAN~\cite{zhang2022efficient} &
   &
  \multicolumn{1}{c|}{32.75} &
  0.9022 &
  \multicolumn{1}{c|}{28.96} &
  0.7914 &
  \multicolumn{1}{c|}{27.83} &
  0.7459 &
  \multicolumn{1}{c|}{27.13} &
  0.8167 &
  \multicolumn{1}{c|}{31.68} &
  0.9226 \\
IPT$^\ast$~\cite{chen2021pre} &
   &
  \multicolumn{1}{c|}{32.64} &
  - &
  \multicolumn{1}{c|}{29.01} &
  - &
  \multicolumn{1}{c|}{27.82} &
  - &
  \multicolumn{1}{c|}{27.26} &
  - &
  \multicolumn{1}{c|}{-} &
  - \\
RCAN-it~\cite{lin2022revisiting} &
   &
  \multicolumn{1}{c|}{32.69} &
  0.9007 &
  \multicolumn{1}{c|}{28.99} &
  0.7922 &
  \multicolumn{1}{c|}{27.87} &
  0.7459 &
  \multicolumn{1}{c|}{27.16} &
  0.8168 &
  \multicolumn{1}{c|}{31.78} &
  0.9217 \\
SwinIR~\cite{liang2021swinir} &
   &
  \multicolumn{1}{c|}{32.92} &
  0.9044 &
  \multicolumn{1}{c|}{29.09} &
  0.7950 &
  \multicolumn{1}{c|}{27.92} &
  0.7489 &
  \multicolumn{1}{c|}{27.45} &
  0.8254 &
  \multicolumn{1}{c|}{32.03} &
  0.9260 \\
CAT-A~\cite{chen2022cross} &
   &
  \multicolumn{1}{c|}{33.08} &
  0.9052 &
  \multicolumn{1}{c|}{29.18} &
  0.7960 &
  \multicolumn{1}{c|}{27.99} &
  0.7510 &
  \multicolumn{1}{c|}{27.89} &
  0.8339 &
  \multicolumn{1}{c|}{32.39} &
  0.9285 \\
DAT~\cite{chen2023dual} &
   &
  \multicolumn{1}{c|}{\color{green}33.08} &
  0.9055 &
  \multicolumn{1}{c|}{\color{green}29.23} &
  \color{green}0.7973 &
  \multicolumn{1}{c|}{\color{green}28.00} &
  0.7515 &
  \multicolumn{1}{c|}{27.87} &
  0.8343 &
  \multicolumn{1}{c|}{\color{green}32.51} &
  0.9291 \\
HAT~\cite{chen2023activating} &
   &
  \multicolumn{1}{c|}{33.04} &
  \color{green}0.9056 &
  \multicolumn{1}{c|}{\color{green}29.23} &
  \color{green}0.7973 &
  \multicolumn{1}{c|}{\color{green}28.00} &
  \color{green}0.7517 &
  \multicolumn{1}{c|}{\color{green}27.97} &
  \color{green}0.8368 &
  \multicolumn{1}{c|}{32.48} &
  \color{green}0.9292 \\
\textbf{CPAT (Ours)} &
   &
  \multicolumn{1}{c|}{\color{blue}33.19} &
  \color{blue}0.9069 &
  \multicolumn{1}{c|}{\color{blue}29.34} &
  \color{blue}0.7991 &
  \multicolumn{1}{c|}{\color{blue}28.04} &
  \color{blue}0.7527 &
  \multicolumn{1}{c|}{\color{blue}28.22} &
  \color{blue}0.8408 &
  \multicolumn{1}{c|}{\color{blue}32.69} &
  \color{blue}0.9309 \\
\textbf{CPAT† (Ours)} &
   &
  \multicolumn{1}{c|}{\color{red}33.24} &
  \color{red}0.9071 &
  \multicolumn{1}{c|}{\color{red}29.36} &
  \color{red}0.7996 &
  \multicolumn{1}{c|}{\color{red}28.06} &
  \color{red}0.7532 &
  \multicolumn{1}{c|}{\color{red}28.33} &
  \color{red}0.8425 &
  \multicolumn{1}{c|}{\color{red}32.85} &
  \color{red}0.9318 \\ \hline
\end{tabular}}
\vspace{3.0mm}
\caption{Quantitative comparison with state-of-the-art methods. The best, second-best, and third-best results are marked in \textcolor{red}{red}, \textcolor{blue}{blue}, and \textcolor{green}{green} colors, respectively. ``$\dagger$” indicates that self-ensemble is used. IPT$^\ast$~\cite{chen2021pre} is trained on ImageNet.}
\label{results_table_1}
\vspace{-4mm}
\end{table}

\begin{figure}[h!]
\begin{center}
\scalebox{1}{
\includegraphics[width=1.0\textwidth]{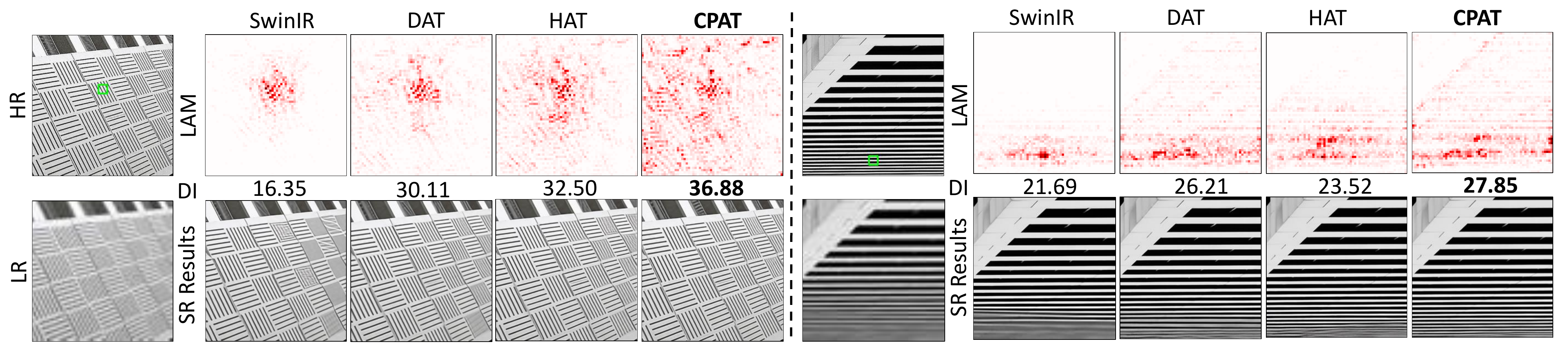}}
\vspace{1mm}
\caption{LAM~\cite{gu2021interpreting} and DI~\cite{gu2021interpreting} comparison results.}\label{fig:LAM_visual}
\end{center}
\vspace{-4mm}
\end{figure}

\vspace{0.0mm}
\subsection{Comparison with State-of-the-Art Methods}

\noindent
\textbf{Quantitative results.} Tab.~\ref{results_table_1} reports the quantitative comparison of CPAT with different state-of-the-art methods including EDSR~\cite{lim2017enhanced}, RCAN~\cite{zhang2018image}, NLSA~\cite{9578003}, ELAN~\cite{zhang2022efficient}, RCAN-it~\cite{lin2022revisiting}, SwinIR~\cite{liang2021swinir}, CAT-A~\cite{chen2022cross}, DAT~\cite{chen2023dual}, and HAT~\cite{chen2023activating}. Our method surpasses the current methods on all benchmark datasets with all scales. The highest increase achieved is 0.31dB on Urban100 on x2 SR when compared with HAT. CPAT improves by more than 0.7dB compared to SwinIR, which uses Swin Transformer as its central backbone for all scales. With the self-ensemble strategy in testing, CPAT† performs better than CPAT, but the inference time is much longer and not helpful for high-resolution input images. All quantitative results demonstrate that enhancing the windows along the height and width of feature maps instead of using the squared windows when computing attention in CPWin-SA and leveraging frequency features in SFIM are very effective for improving the quality of the SR image.

\begin{figure}[t!]
\begin{center}
\scalebox{0.95}{
\includegraphics[width=1.0\textwidth]{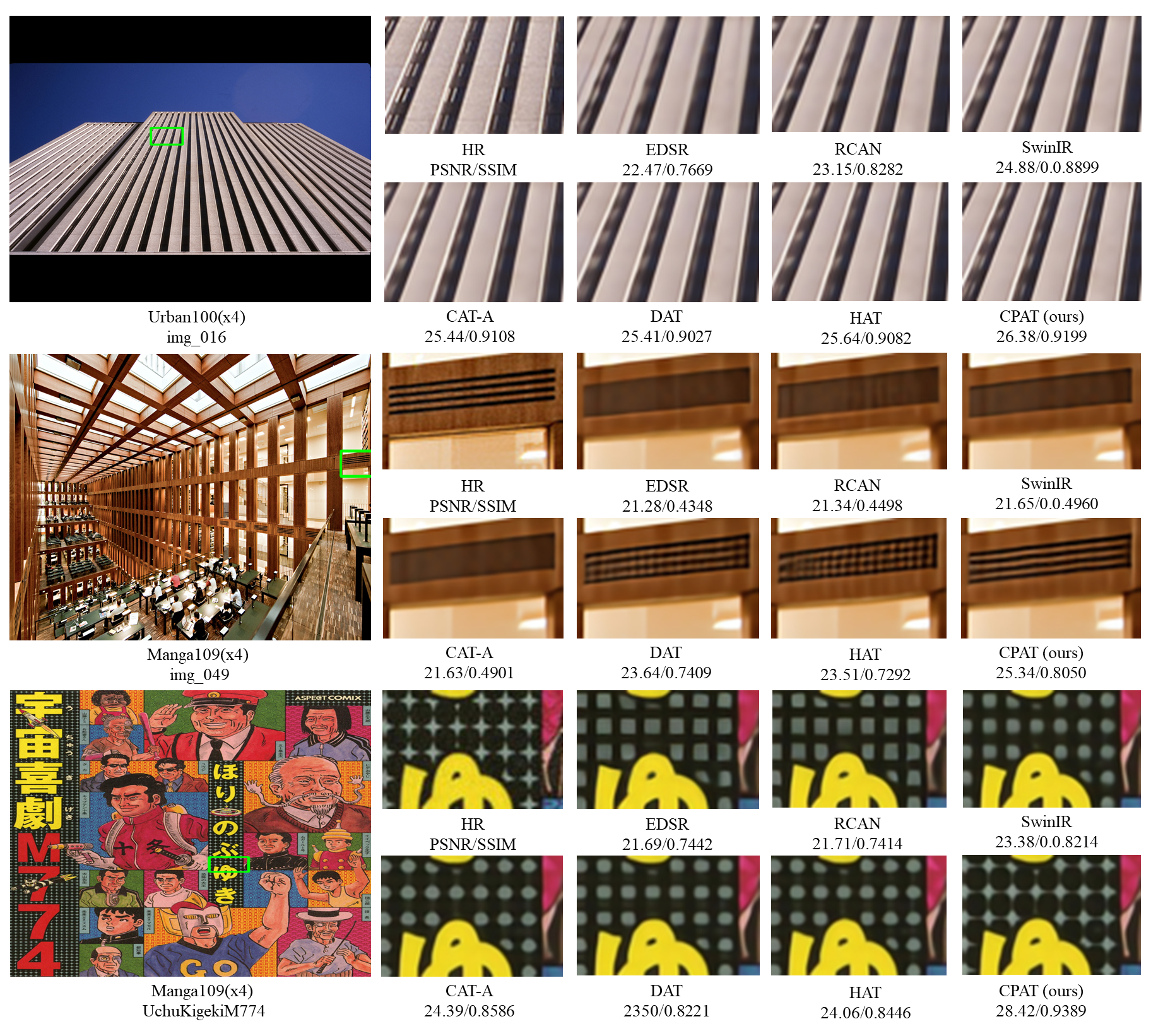}}\\
\vspace{2mm}
\caption{Qualitative comparison (x4 SR). The patch images being compared are the green boxes in the HR images. PSNR/SSIM is also computed correspondingly on these patches to demonstrate the improvement of our method.}
\label{fig:qualitative_result}
\end{center}
\vspace{-9.0mm}
\end{figure}

\noindent
\textbf{Qualitative results.} Local Attribution Map (LAM)~\cite{gu2021interpreting} and Diffusion Index (DI)~\cite{gu2021interpreting} comparisons are shown in Fig.~\ref{fig:LAM_visual}.
LAM emphasizes the significance of pixels in the LR image during upscaling of the patches marked with green boxes. DI is indicative of the wider range of pixels utilized. A higher DI indicates a wider range of pixels in upscaling images. LAM and DI results show the superiority of our method over other methods. The visual comparison is shown in Fig.~\ref{fig:qualitative_result} with $"img\_16"$, $"img\_49"$ from Urban100, and $"UchuKigekiM774"$ from Manga109. Our method can enhance the details of the LR image more clearly, with less blur, and higher PSNR/SSIM compared to other methods. All qualitative results show the effectiveness of our method for SISR. More details on qualitative results, self-ensemble strategy, and lightweight version of CPAT can be found in the $supp.$ file.

\subsection{Model Complexity Analysis}

\begin{wraptable}{r}{0.39\textwidth}
\vspace{-4mm}
\setlength{\tabcolsep}{1.5mm}
\scalebox{0.82}{
\begin{tabular}[t]{|c|c|c|c|c|c|}
\hline
Models & EDSR  & HAT & CPAT\\ \hline\hline
Params & 40.73M  & 20.62M & 20.39M\\
\hline
FLOPs & 667.92G & 334.09G & 329.04G\\
\hline
PSNR & 26.64 & 27.97 & \textbf{28.21}\\
\hline
\end{tabular}}
\vspace{3mm}
\caption{Model Complexity Comparison EDSR, HAT and CPAT.}\label{wrap-tab:15}
\vspace{-2mm}
\end{wraptable}

We analyze the computational complexity of our method with the CNN-based method (EDSR) and the Transformer-based method (HAT). The x4 SR performance on Urban100 and FLOPs are computed at the output SR image size of $256\times 256$, reported in Tab.~\ref{wrap-tab:15}.
\noindent
We can make the following observations: 1) CPAT performs better both in terms of performance and model complexity compared to EDSR. Specifically, the number of parameters and FLOPs of EDSR are both about twice as high as CPAT's, while the PSNR is significantly lower. 2) CPAT still performs better than the current SOTA method HAT. Specifically, HAT achieves 27.97dB on Urban100 for x2 SR, while CPAT is 28.21dB. In terms of model complexity, CPAT is also slightly better than HAT in both the number of parameters and number of FLOPs. Specifically, CPAT has 20.39M parameters compared to HAT's 20.62M. Similarly, CPAT has a number of FLOPs of 329.04G, while HAT has 334.09G. The results show that our method achieves significant improvement while still ensuring computational cost compared to current methods.

\vspace{-4mm}
\section{Conclusions}
\label{sec:conclusion}
\vspace{-2.0mm}
In this study, we propose a novel Channel-Partitioned Attention Transformer (CPAT) for SISR. This Transformer consists of V-EWin, H-EWin, and W-MSA attentions. EWin and H-EWin enhance windows along the height and width of input features to better capture long-range dependencies and relationships between distant tokens. We also use squared window-based attention in CPAT, which focuses on local features. We calculate self-attention in various window shapes to apply to datasets that contain texture features in various directions (\eg., Urban100~\cite{7299156}). Additionally, we propose Spatial-Frequency Interaction Module (SFIM), which is simple yet effectively leverages features (patterns, textures, edges, $etc.$) from the frequency domain that might be less apparent in the spatial domain. Integrating frequency spatial features helps to achieve comprehensive information from feature maps that is important for HR image reconstruction. Based on the proposals above, our method outperforms the current methods in both quantitative and qualitative results.

\section*{Acknowledgments}
\label{sec:acknowledgments}
This work was supported by the Institute for Information \& communications Technology Promotion (IITP) grant funded by the Korea government (MSIP) (No. 2022-0-00407); and the MSIT(Ministry of Science and ICT), Korea, under the Grand Information Technology Research Center support program(IITP-2022-2020-0-01489) supervised by the IITP(Institute for Information \& communications Technology Planning \& Evaluation)

\bibliography{egbib}

\begin{thebibliography}{58}
\providecommand{\natexlab}[1]{#1}
\providecommand{\url}[1]{\texttt{#1}}
\expandafter\ifx\csname urlstyle\endcsname\relax
  \providecommand{\doi}[1]{doi: #1}\else
  \providecommand{\doi}{doi: \begingroup \urlstyle{rm}\Url}\fi

\bibitem[Agustsson and Timofte(2017)]{8014884}
E.~Agustsson and R.~Timofte.
\newblock Ntire 2017 challenge on single image super-resolution: Dataset and study.
\newblock In \emph{2017 IEEE Conference on Computer Vision and Pattern Recognition Workshops (CVPRW)}, pages 1122--1131, Los Alamitos, CA, USA, jul 2017. IEEE Computer Society.
\newblock \doi{10.1109/CVPRW.2017.150}.
\newblock URL \url{https://doi.ieeecomputersociety.org/10.1109/CVPRW.2017.150}.

\bibitem[Ahmed et~al.(1974)Ahmed, Natarajan, and Rao]{1672377}
N.~Ahmed, T.~Natarajan, and K.R. Rao.
\newblock Discrete cosine transform.
\newblock \emph{IEEE Transactions on Computers}, C-23\penalty0 (1):\penalty0 90--93, 1974.
\newblock \doi{10.1109/T-C.1974.223784}.

\bibitem[Bevilacqua et~al.(2012)Bevilacqua, Roumy, Guillemot, and Alberi-Morel]{Bevilacqua2012LowComplexitySS}
Marco Bevilacqua, Aline Roumy, Christine~M. Guillemot, and Marie-Line Alberi-Morel.
\newblock Low-complexity single-image super-resolution based on nonnegative neighbor embedding.
\newblock In \emph{British Machine Vision Conference}, 2012.
\newblock URL \url{https://api.semanticscholar.org/CorpusID:5250573}.

\bibitem[Bluche(2016)]{bluche2016joint}
Th{\'e}odore Bluche.
\newblock Joint line segmentation and transcription for end-to-end handwritten paragraph recognition.
\newblock \emph{Advances in neural information processing systems}, 29, 2016.

\bibitem[Brigham(1988)]{brigham1988fast}
E~Oran Brigham.
\newblock \emph{The fast Fourier transform and its applications}.
\newblock Prentice-Hall, Inc., 1988.

\bibitem[Cai et~al.(2021)Cai, Ding, and Lu]{Cai2021FreqNetAF}
Runyuan Cai, Yue Ding, and Hongtao Lu.
\newblock Freqnet: A frequency-domain image super-resolution network with dicrete cosine transform.
\newblock \emph{ArXiv}, abs/2111.10800, 2021.
\newblock URL \url{https://api.semanticscholar.org/CorpusID:244478402}.

\bibitem[Chen et~al.(2021{\natexlab{a}})Chen, Wang, Guo, Xu, Deng, Liu, Ma, Xu, Xu, and Gao]{chen2021pre}
Hanting Chen, Yunhe Wang, Tianyu Guo, Chang Xu, Yiping Deng, Zhenhua Liu, Siwei Ma, Chunjing Xu, Chao Xu, and Wen Gao.
\newblock Pre-trained image processing transformer.
\newblock In \emph{Proceedings of the IEEE/CVF conference on computer vision and pattern recognition}, pages 12299--12310, 2021{\natexlab{a}}.

\bibitem[Chen et~al.(2021{\natexlab{b}})Chen, Lu, Yu, Luo, Adeli, Wang, Lu, Yuille, and Zhou]{chen2021transunet}
Jieneng Chen, Yongyi Lu, Qihang Yu, Xiangde Luo, Ehsan Adeli, Yan Wang, Le~Lu, Alan~L Yuille, and Yuyin Zhou.
\newblock Transunet: Transformers make strong encoders for medical image segmentation.
\newblock \emph{arXiv preprint arXiv:2102.04306}, 2021{\natexlab{b}}.

\bibitem[Chen et~al.(2022{\natexlab{a}})Chen, Du, Yang, Beyer, Zhai, Lin, Chen, Li, Song, Wang, and Zhou]{10.1007/978-3-031-20080-9_41}
Wuyang Chen, Xianzhi Du, Fan Yang, Lucas Beyer, Xiaohua Zhai, Tsung-Yi Lin, Huizhong Chen, Jing Li, Xiaodan Song, Zhangyang Wang, and Denny Zhou.
\newblock A simple single-scale vision transformer for object detection and instance segmentation.
\newblock In \emph{Computer Vision – ECCV 2022: 17th European Conference, Tel Aviv, Israel, October 23–27, 2022, Proceedings, Part X}, page 711–727, Berlin, Heidelberg, 2022{\natexlab{a}}. Springer-Verlag.
\newblock ISBN 978-3-031-20079-3.
\newblock \doi{10.1007/978-3-031-20080-9_41}.
\newblock URL \url{https://doi.org/10.1007/978-3-031-20080-9_41}.

\bibitem[Chen et~al.(2023{\natexlab{a}})Chen, Wang, Zhou, Qiao, and Dong]{chen2023activating}
Xiangyu Chen, Xintao Wang, Jiantao Zhou, Yu~Qiao, and Chao Dong.
\newblock Activating more pixels in image super-resolution transformer.
\newblock In \emph{Proceedings of the IEEE/CVF conference on computer vision and pattern recognition}, pages 22367--22377, 2023{\natexlab{a}}.

\bibitem[Chen et~al.(2022{\natexlab{b}})Chen, Zhang, Gu, Kong, Yuan, et~al.]{chen2022cross}
Zheng Chen, Yulun Zhang, Jinjin Gu, Linghe Kong, Xin Yuan, et~al.
\newblock Cross aggregation transformer for image restoration.
\newblock \emph{Advances in Neural Information Processing Systems}, 35:\penalty0 25478--25490, 2022{\natexlab{b}}.

\bibitem[Chen et~al.(2023{\natexlab{b}})Chen, Zhang, Gu, Kong, Yang, and Yu]{chen2023dual}
Zheng Chen, Yulun Zhang, Jinjin Gu, Linghe Kong, Xiaokang Yang, and Fisher Yu.
\newblock Dual aggregation transformer for image super-resolution.
\newblock In \emph{Proceedings of the IEEE/CVF international conference on computer vision}, pages 12312--12321, 2023{\natexlab{b}}.

\bibitem[Chu et~al.(2023)Chu, Tian, Zhang, Wang, and Shen]{chu2023CPVT}
Xiangxiang Chu, Zhi Tian, Bo~Zhang, Xinlong Wang, and Chunhua Shen.
\newblock Conditional positional encodings for vision transformers.
\newblock In \emph{ICLR 2023}, 2023.
\newblock URL \url{https://openreview.net/forum?id=3KWnuT-R1bh}.

\bibitem[Cooley and Tukey(1965)]{cooley1965algorithm}
James~W Cooley and John~W Tukey.
\newblock An algorithm for the machine calculation of complex fourier series.
\newblock \emph{Mathematics of computation}, 19\penalty0 (90):\penalty0 297--301, 1965.

\bibitem[Dong et~al.(2015)Dong, Loy, He, and Tang]{dong2015image}
Chao Dong, Chen~Change Loy, Kaiming He, and Xiaoou Tang.
\newblock Image super-resolution using deep convolutional networks.
\newblock \emph{IEEE transactions on pattern analysis and machine intelligence}, 38\penalty0 (2):\penalty0 295--307, 2015.

\bibitem[Dosovitskiy et~al.(2021)Dosovitskiy, Beyer, Kolesnikov, Weissenborn, Zhai, Unterthiner, Dehghani, Minderer, Heigold, Gelly, Uszkoreit, and Houlsby]{dosovitskiy2020vit}
Alexey Dosovitskiy, Lucas Beyer, Alexander Kolesnikov, Dirk Weissenborn, Xiaohua Zhai, Thomas Unterthiner, Mostafa Dehghani, Matthias Minderer, Georg Heigold, Sylvain Gelly, Jakob Uszkoreit, and Neil Houlsby.
\newblock An image is worth 16x16 words: Transformers for image recognition at scale.
\newblock \emph{ICLR}, 2021.

\bibitem[Gohshi(2015)]{7217088}
Seiichi Gohshi.
\newblock Frequency domain analysis for super resolution image reconstruction and its limitations.
\newblock In \emph{2015 10th Asia-Pacific Symposium on Information and Telecommunication Technologies (APSITT)}, pages 1--3, 2015.
\newblock \doi{10.1109/APSITT.2015.7217088}.

\bibitem[Grechishnikova(2019)]{Grechishnikova863415}
Daria Grechishnikova.
\newblock Transformer neural network for protein specific de novo drug generation as machine translation problem.
\newblock \emph{bioRxiv}, 2019.
\newblock \doi{10.1101/863415}.
\newblock URL \url{https://www.biorxiv.org/content/early/2019/12/03/863415}.

\bibitem[Gu and Dong(2021)]{gu2021interpreting}
Jinjin Gu and Chao Dong.
\newblock Interpreting super-resolution networks with local attribution maps.
\newblock In \emph{Proceedings of the IEEE/CVF Conference on Computer Vision and Pattern Recognition}, pages 9199--9208, 2021.

\bibitem[He et~al.(2016)He, Chen, and Liu]{he2016salient}
Chao He, Zhenxue Chen, and Chengyun Liu.
\newblock Salient object detection via images frequency domain analyzing.
\newblock \emph{Signal, Image and Video Processing}, 10:\penalty0 1295--1302, 2016.

\bibitem[He et~al.(2020)He, Liao, Tavakoli, Yang, Rosenhahn, and Pugeault]{he2020image}
Sen He, Wentong Liao, Hamed~R Tavakoli, Michael Yang, Bodo Rosenhahn, and Nicolas Pugeault.
\newblock Image captioning through image transformer.
\newblock In \emph{Proceedings of the Asian conference on computer vision}, 2020.

\bibitem[Hendrycks and Gimpel(2016)]{hendrycks2016gelu}
Dan Hendrycks and Kevin Gimpel.
\newblock Gaussian error linear units (gelus).
\newblock \emph{arXiv preprint arXiv:1606.08415}, 2016.

\bibitem[Huang et~al.(2015)Huang, Singh, and Ahuja]{7299156}
Jia-Bin Huang, Abhishek Singh, and Narendra Ahuja.
\newblock Single image super-resolution from transformed self-exemplars.
\newblock In \emph{2015 IEEE Conference on Computer Vision and Pattern Recognition (CVPR)}, pages 5197--5206, 2015.
\newblock \doi{10.1109/CVPR.2015.7299156}.

\bibitem[Huang et~al.(2023)Huang, Wang, Wei, Huang, Shi, Liu, and Huang]{9133304}
Z.~Huang, X.~Wang, Y.~Wei, L.~Huang, H.~Shi, W.~Liu, and T.~S. Huang.
\newblock Ccnet: Criss-cross attention for semantic segmentation.
\newblock \emph{IEEE Transactions on Pattern Analysis \& Machine Intelligence}, 45\penalty0 (06):\penalty0 6896--6908, jun 2023.
\newblock ISSN 1939-3539.
\newblock \doi{10.1109/TPAMI.2020.3007032}.

\bibitem[Jaderberg et~al.(2015)Jaderberg, Simonyan, Zisserman, et~al.]{jaderberg2015spatial}
Max Jaderberg, Karen Simonyan, Andrew Zisserman, et~al.
\newblock Spatial transformer networks.
\newblock \emph{Advances in neural information processing systems}, 28, 2015.

\bibitem[Kim et~al.(2016{\natexlab{a}})Kim, Lee, and Lee]{kim2016accurate}
Jiwon Kim, Jung~Kwon Lee, and Kyoung~Mu Lee.
\newblock Accurate image super-resolution using very deep convolutional networks.
\newblock In \emph{Proceedings of the IEEE conference on computer vision and pattern recognition}, pages 1646--1654, 2016{\natexlab{a}}.

\bibitem[Kim et~al.(2016{\natexlab{b}})Kim, Lee, and Lee]{kim2016deeply}
Jiwon Kim, Jung~Kwon Lee, and Kyoung~Mu Lee.
\newblock Deeply-recursive convolutional network for image super-resolution.
\newblock In \emph{Proceedings of the IEEE conference on computer vision and pattern recognition}, pages 1637--1645, 2016{\natexlab{b}}.

\bibitem[Kim et~al.(2001)Kim, Kim, Lee, and Lim]{kim2001signal}
JM~Kim, SH~Kim, DJ~Lee, and HS~Lim.
\newblock Signal processing using fourier \& wavelet transform for pulse oximetry.
\newblock In \emph{Technical Digest. CLEO/Pacific Rim 2001. 4th Pacific Rim Conference on Lasers and Electro-Optics (Cat. No. 01TH8557)}, volume~2, pages II--II. IEEE, 2001.

\bibitem[Kingma and Ba(2014)]{Kingma2014AdamAM}
Diederik~P. Kingma and Jimmy Ba.
\newblock Adam: A method for stochastic optimization.
\newblock \emph{CoRR}, abs/1412.6980, 2014.
\newblock URL \url{https://api.semanticscholar.org/CorpusID:6628106}.

\bibitem[Ledig et~al.(2017)Ledig, Theis, Husz{\'a}r, Caballero, Cunningham, Acosta, Aitken, Tejani, Totz, Wang, et~al.]{ledig2017photo}
Christian Ledig, Lucas Theis, Ferenc Husz{\'a}r, Jose Caballero, Andrew Cunningham, Alejandro Acosta, Andrew Aitken, Alykhan Tejani, Johannes Totz, Zehan Wang, et~al.
\newblock Photo-realistic single image super-resolution using a generative adversarial network.
\newblock In \emph{Proceedings of the IEEE conference on computer vision and pattern recognition}, pages 4681--4690, 2017.

\bibitem[Li et~al.(2018)Li, Fang, Mei, and Zhang]{Li_2018_ECCV}
Juncheng Li, Faming Fang, Kangfu Mei, and Guixu Zhang.
\newblock Multi-scale residual network for image super-resolution.
\newblock In \emph{Proceedings of the European Conference on Computer Vision (ECCV)}, September 2018.

\bibitem[Liang et~al.(2021)Liang, Cao, Sun, Zhang, Van~Gool, and Timofte]{liang2021swinir}
Jingyun Liang, Jiezhang Cao, Guolei Sun, Kai Zhang, Luc Van~Gool, and Radu Timofte.
\newblock Swinir: Image restoration using swin transformer.
\newblock In \emph{Proceedings of the IEEE/CVF international conference on computer vision}, pages 1833--1844, 2021.

\bibitem[Lim et~al.(2017)Lim, Son, Kim, Nah, and Mu~Lee]{lim2017enhanced}
Bee Lim, Sanghyun Son, Heewon Kim, Seungjun Nah, and Kyoung Mu~Lee.
\newblock Enhanced deep residual networks for single image super-resolution.
\newblock In \emph{Proceedings of the IEEE conference on computer vision and pattern recognition workshops}, pages 136--144, 2017.

\bibitem[Lin et~al.(2022)Lin, Garg, Banerjee, Magid, Sun, Zhang, Van~Gool, Wei, and Pfister]{lin2022revisiting}
Zudi Lin, Prateek Garg, Atmadeep Banerjee, Salma~Abdel Magid, Deqing Sun, Yulun Zhang, Luc Van~Gool, Donglai Wei, and Hanspeter Pfister.
\newblock Revisiting rcan: Improved training for image super-resolution.
\newblock \emph{arXiv preprint arXiv:2201.11279}, 2022.

\bibitem[Liu et~al.(2021)Liu, Lin, Cao, Hu, Wei, Zhang, Lin, and Guo]{liu2021swin}
Ze~Liu, Yutong Lin, Yue Cao, Han Hu, Yixuan Wei, Zheng Zhang, Stephen Lin, and Baining Guo.
\newblock Swin transformer: Hierarchical vision transformer using shifted windows.
\newblock In \emph{Proceedings of the IEEE/CVF international conference on computer vision}, pages 10012--10022, 2021.

\bibitem[Maas et~al.(2013)Maas, Hannun, Ng, et~al.]{maas2013rectifier}
Andrew~L Maas, Awni~Y Hannun, Andrew~Y Ng, et~al.
\newblock Rectifier nonlinearities improve neural network acoustic models.
\newblock In \emph{Proc. icml}, volume~30, page~3. Atlanta, GA, 2013.

\bibitem[Martin et~al.(2001)Martin, Fowlkes, Tal, and Malik]{937655}
D.~Martin, C.~Fowlkes, D.~Tal, and J.~Malik.
\newblock A database of human segmented natural images and its application to evaluating segmentation algorithms and measuring ecological statistics.
\newblock In \emph{Proceedings Eighth IEEE International Conference on Computer Vision. ICCV 2001}, volume~2, pages 416--423 vol.2, 2001.
\newblock \doi{10.1109/ICCV.2001.937655}.

\bibitem[Matsui et~al.(2016)Matsui, Ito, Aramaki, Fujimoto, Ogawa, Yamasaki, and Aizawa]{Matsui_2016}
Yusuke Matsui, Kota Ito, Yuji Aramaki, Azuma Fujimoto, Toru Ogawa, Toshihiko Yamasaki, and Kiyoharu Aizawa.
\newblock Sketch-based manga retrieval using manga109 dataset.
\newblock \emph{Multimedia Tools and Applications}, 76\penalty0 (20):\penalty0 21811–21838, November 2016.
\newblock ISSN 1573-7721.
\newblock \doi{10.1007/s11042-016-4020-z}.
\newblock URL \url{http://dx.doi.org/10.1007/s11042-016-4020-z}.

\bibitem[Mei et~al.(2021)Mei, Fan, and Zhou]{9578003}
Yiqun Mei, Yuchen Fan, and Yuqian Zhou.
\newblock Image super-resolution with non-local sparse attention.
\newblock In \emph{2021 IEEE/CVF Conference on Computer Vision and Pattern Recognition (CVPR)}, pages 3516--3525, 2021.
\newblock \doi{10.1109/CVPR46437.2021.00352}.

\bibitem[Miech et~al.(2017)Miech, Laptev, and Sivic]{miech17loupe}
Antoine Miech, Ivan Laptev, and Josef Sivic.
\newblock Learnable pooling with context gating for video classification.
\newblock \emph{arXiv:1706.06905}, 2017.

\bibitem[Rambhatla et~al.(2023)Rambhatla, Misra, Chellappa, and Shrivastava]{rambhatla2023most}
Sai~Saketh Rambhatla, Ishan Misra, Rama Chellappa, and Abhinav Shrivastava.
\newblock Most: Multiple object localization with self-supervised transformers for object discovery.
\newblock In \emph{Proceedings of the IEEE/CVF International Conference on Computer Vision}, pages 15823--15834, 2023.

\bibitem[Sargent et~al.(2023)Sargent, Koh, Zhang, Chang, Herrmann, Srinivasan, Wu, and Sun]{sargent2023vq3d}
Kyle Sargent, Jing~Yu Koh, Han Zhang, Huiwen Chang, Charles Herrmann, Pratul Srinivasan, Jiajun Wu, and Deqing Sun.
\newblock Vq3d: Learning a 3d-aware generative model on imagenet.
\newblock In \emph{Proceedings of the IEEE/CVF International Conference on Computer Vision}, pages 4240--4250, 2023.

\bibitem[Saxena and Singh(2005)]{saxena2005fractional}
Rajiv Saxena and Kulbir Singh.
\newblock Fractional fourier transform: A novel tool for signal processing.
\newblock \emph{Journal of the Indian Institute of Science}, 85\penalty0 (1):\penalty0 11, 2005.

\bibitem[Shi et~al.(2022)Shi, Jiang, Dai, and Schiele]{shi2022motion}
Shaoshuai Shi, Li~Jiang, Dengxin Dai, and Bernt Schiele.
\newblock Motion transformer with global intention localization and local movement refinement.
\newblock \emph{Advances in Neural Information Processing Systems}, 35:\penalty0 6531--6543, 2022.

\bibitem[Shi et~al.(2016)Shi, Caballero, Huszar, Totz, Aitken, Bishop, Rueckert, and Wang]{7780576}
W.~Shi, J.~Caballero, F.~Huszar, J.~Totz, A.~P. Aitken, R.~Bishop, D.~Rueckert, and Z.~Wang.
\newblock Real-time single image and video super-resolution using an efficient sub-pixel convolutional neural network.
\newblock In \emph{2016 IEEE Conference on Computer Vision and Pattern Recognition (CVPR)}, pages 1874--1883, Los Alamitos, CA, USA, jun 2016. IEEE Computer Society.
\newblock \doi{10.1109/CVPR.2016.207}.
\newblock URL \url{https://doi.ieeecomputersociety.org/10.1109/CVPR.2016.207}.

\bibitem[Sun et~al.(2022)Sun, Zhou, Black, and Chandrasekaran]{sun2022locate}
Jiankai Sun, Bolei Zhou, Michael~J Black, and Arjun Chandrasekaran.
\newblock Locate: End-to-end localization of actions in 3d with transformers.
\newblock \emph{arXiv preprint arXiv:2203.10719}, 2022.

\bibitem[Timofte et~al.(2016)Timofte, Rothe, and Van~Gool]{timofte2016seven}
Radu Timofte, Rasmus Rothe, and Luc Van~Gool.
\newblock Seven ways to improve example-based single image super resolution.
\newblock In \emph{Proceedings of the IEEE conference on computer vision and pattern recognition}, pages 1865--1873, 2016.

\bibitem[Tran et~al.(2022)Tran, Nguyen, Pham, and Tran]{TranNPT22}
Dinh{-}Phu Tran, Quoc{-}Anh Nguyen, Van{-}Truong Pham, and Thi{-}Thao Tran.
\newblock Trans2unet: Neural fusion for nuclei semantic segmentation.
\newblock In \emph{11th International Conference on Control, Automation and Information Sciences, {ICCAIS} 2022, Hanoi, Vietnam, November 21-24, 2022}, pages 583--588. {IEEE}, 2022.
\newblock \doi{10.1109/ICCAIS56082.2022.9990159}.
\newblock URL \url{https://doi.org/10.1109/ICCAIS56082.2022.9990159}.

\bibitem[Trider(1978)]{trider1978fast}
R~Trider.
\newblock A fast fourier transform (fft) based sonar signal processor.
\newblock \emph{IEEE Transactions on Acoustics, Speech, and Signal Processing}, 26\penalty0 (1):\penalty0 15--20, 1978.

\bibitem[Wang et~al.(2023)Wang, Jiang, Zhong, and Liu]{Wang_2023_CVPR}
Chenyang Wang, Junjun Jiang, Zhiwei Zhong, and Xianming Liu.
\newblock Spatial-frequency mutual learning for face super-resolution.
\newblock In \emph{Proceedings of the IEEE/CVF Conference on Computer Vision and Pattern Recognition (CVPR)}, pages 22356--22366, June 2023.

\bibitem[Wang et~al.(2022)Wang, Xu, and Sun]{wang2022end}
Yiyu Wang, Jungang Xu, and Yingfei Sun.
\newblock End-to-end transformer based model for image captioning.
\newblock In \emph{Proceedings of the AAAI Conference on Artificial Intelligence}, volume~36, pages 2585--2594, 2022.

\bibitem[Wang et~al.(2004)Wang, Bovik, Sheikh, and Simoncelli]{wang2004image}
Zhou Wang, Alan~C Bovik, Hamid~R Sheikh, and Eero~P Simoncelli.
\newblock Image quality assessment: from error visibility to structural similarity.
\newblock \emph{IEEE transactions on image processing}, 13\penalty0 (4):\penalty0 600--612, 2004.

\bibitem[Zeyde et~al.(2010)Zeyde, Elad, and Protter]{inproceedings}
Roman Zeyde, Michael Elad, and Matan Protter.
\newblock On single image scale-up using sparse-representations.
\newblock volume 6920, pages 711--730, 06 2010.
\newblock ISBN 978-3-642-27412-1.
\newblock \doi{10.1007/978-3-642-27413-8_47}.

\bibitem[Zhang et~al.(2022)Zhang, Zeng, Guo, and Zhang]{zhang2022efficient}
Xindong Zhang, Hui Zeng, Shi Guo, and Lei Zhang.
\newblock Efficient long-range attention network for image super-resolution.
\newblock In \emph{European conference on computer vision}, pages 649--667. Springer, 2022.

\bibitem[Zhang et~al.(2018{\natexlab{a}})Zhang, Li, Li, Wang, Zhong, and Fu]{zhang2018image}
Yulun Zhang, Kunpeng Li, Kai Li, Lichen Wang, Bineng Zhong, and Yun Fu.
\newblock Image super-resolution using very deep residual channel attention networks.
\newblock In \emph{Proceedings of the European conference on computer vision (ECCV)}, pages 286--301, 2018{\natexlab{a}}.

\bibitem[Zhang et~al.(2018{\natexlab{b}})Zhang, Tian, Kong, Zhong, and Fu]{zhang2018residual}
Yulun Zhang, Yapeng Tian, Yu~Kong, Bineng Zhong, and Yun Fu.
\newblock Residual dense network for image super-resolution.
\newblock In \emph{Proceedings of the IEEE conference on computer vision and pattern recognition}, pages 2472--2481, 2018{\natexlab{b}}.

\bibitem[Zhao et~al.(2018)Zhao, Li, Zhao, and Feng]{zhao2018weakly}
Fang Zhao, Jianshu Li, Jian Zhao, and Jiashi Feng.
\newblock Weakly supervised phrase localization with multi-scale anchored transformer network.
\newblock In \emph{Proceedings of the IEEE Conference on Computer Vision and Pattern Recognition}, pages 5696--5705, 2018.

\bibitem[Zhu et~al.(2022)Zhu, Shah, and Chen]{zhu2022transgeo}
Sijie Zhu, Mubarak Shah, and Chen Chen.
\newblock Transgeo: Transformer is all you need for cross-view image geo-localization.
\newblock In \emph{Proceedings of the IEEE/CVF Conference on Computer Vision and Pattern Recognition}, pages 1162--1171, 2022.

\end{thebibliography}
\end{document}